\documentclass[10pt,twocolumn,letterpaper]{article}

\usepackage[pagenumbers]{cvpr} % To force page numbers, e.g. for an arXiv version

\definecolor{cvprblue}{rgb}{0.21,0.49,0.74}
\usepackage[pagebackref,breaklinks,colorlinks,allcolors=cvprblue]{hyperref}
\usepackage{bbm}
\usepackage{algorithm}
\usepackage{algpseudocode}
\usepackage{listings}
\usepackage[frozencache,cachedir=minted-cache]{minted}
\usepackage{float} 
\usepackage{pifont}
\usepackage{hyperref}
\def\paperID{} % *** Enter the Paper ID here
\def\confName{CVPR}
\def\confYear{2026}

\title{Splatwizard: A Benchmark Toolkit for 3D Gaussian Splatting Compression}

\author{
Xiang Liu\textsuperscript{1,3}\quad
Yimin Zhou\textsuperscript{1} \quad
Jinxiang Wang\textsuperscript{1} \quad
Yujun Huang\textsuperscript{1} \quad
Shuzhao Xie\textsuperscript{1} \quad
Shiyu Qin\textsuperscript{1}\\
Mingyao Hong\textsuperscript{3}  \quad
Jiawei Li\textsuperscript{4}  \quad
Yaowei Wang\textsuperscript{2,3}  \quad
Zhi Wang\textsuperscript{1}  \quad
Shu-Tao Xia\textsuperscript{1}  \quad
Bin Chen\textsuperscript{2,}\thanks{Corresponding author}\\
\textsuperscript{1}Tsinghua University 
\textsuperscript{2}Harbin Institute of Technology, Shenzhen\\
\textsuperscript{3}Pengcheng Laboratory 
\textsuperscript{4}Huawei Technologies Ltd.
}

\begin{document}
\maketitle
\begin{abstract}
The recent advent of 3D Gaussian Splatting (3DGS) has marked a significant breakthrough in real-time novel view synthesis. However, the rapid proliferation of 3DGS-based algorithms has created a pressing need for standardized and comprehensive evaluation tools, especially for compression task. Existing benchmarks often lack the specific metrics necessary to holistically assess the unique characteristics of different methods, such as rendering speed, rate distortion trade-offs memory efficiency, and geometric accuracy. To address this gap, we introduce Splatwizard, a unified benchmark toolkit designed specifically for benchmarking 3DGS compression models. Splatwizard provides an easy-to-use framework to implement new 3DGS compression model and utilize state-of-the-art techniques proposed by previous work. Besides, an integrated pipeline that automates the calculation of key performance indicators, including image-based quality metrics, chamfer distance of reconstruct mesh, rendering frame rates, and computational resource consumption is included in the framework as well. Code is available at \href{https://github.com/splatwizard/splatwizard}{Github}.

\end{abstract}    
\section{Introduction}
\label{sec:intro}

With the rapid progress of computer vision, 3D reconstruction has emerged as a core research area, underpinning downstream applications in virtual reality and embodied intelligence. Among various 3D representations, 3D Gaussian Splatting (3DGS)~\cite{kerbl20233d} achieves state-of-the-art rendering quality and real-time performance, enabling a wide range of practical uses.

\begin{table*}[t]
    \centering
    \begin{tabular}{l c  cc c c c}
    \toprule
    Method                                    & Year  & Std Rasterizer & QAT & Entropy Model & Post Pruning & Compression \\
    \midrule
    3DGS \cite{kerbl20233d}                   & 2023 & \ding{51} & -  & - & \ding{55} & \ding{55}\\
    Trimming the Fat \cite{ali2024trimming}   & 2024 & \ding{51} & -  & - & \ding{51} & \ding{55} \\
    PUP-3DGS \cite{hanson2025pup}             & 2025 & \ding{51} & -  & - & \ding{51} & \ding{55} \\
    Speedy splat \cite{hanson2025speedy}      & 2025 & \ding{51} & -  & - & \ding{55} & \ding{55}\\
    LightGaussian \cite{fan2024lightgaussian} & 2024 & \ding{51} & -  & - & \ding{51} & \ding{51} \\
    MesonGS \cite{xie2024mesongs}             & 2024 & \ding{51} & -  & - & \ding{51} & \ding{51} \\
    % ControlGS \cite{zhang2025consistent}      & 2025 & \ding{51} &    & & &\\
    C3DGS \cite{niedermayr2024compressed}     & 2024 &  \ding{55} & Fake & - & \ding{51} & \ding{51}  \\
    Compact3DGS \cite{lee2024compact}         & 2024 &  \ding{55} & RVQ  &  - & \ding{55} & \ding{51} \\
    HAC \cite{chen2024hac}                    & 2024 &  \ding{55} & Noise  & Gaus      & \ding{55} & \ding{51} \\
    CAT-3DGS \cite{zhan2025cat}               & 2025 &  \ding{55} & Noise  & Gaus/Laplace & \ding{55} & \ding{51} \\
         \bottomrule
    \end{tabular}
    \caption{Comparison of different methods. }
    \label{tab:intro:comparision}
    \vspace{-1.5em}
\end{table*}

However, raw 3DGS scenes contain substantial redundancy. A 3DGS represents a scene with millions of Gaussian ellipsoids distributed in 3D space, whose projected splats jointly capture geometry and appearance. In practice, a single scene often requires up to $10^7$ Gaussians, leading to hundreds of megabytes of storage and limiting the deployability of uncompressed 3DGS in real applications. This has driven the development of numerous 3DGS compression techniques, as summarized in Table~\ref{tab:intro:comparision}. These methods introduce heterogeneous designs across preprocessing, rendering pipelines, and post-processing stages, resulting in large differences in code structure, system behavior, and implementation logic. Consequently, they exhibit notable trade-offs in FPS, memory footprint, model size, and reconstruction quality, making it difficult to achieve consistent evaluation or identify universally effective improvements.

To address this issue, several works~\cite{3DGSzip2025, ye2024gaustudio} have attempted to summarize or benchmark existing 3DGS compression methods. However, these surveys largely rely on reported results or cover only a subset of representative techniques, leaving the practical selection of a suitable method still unclear—especially because downstream tasks prioritize different properties. For instance, streaming systems~\cite{wang2024v3,zhang2022neuvv,10.1145/3570361.3592530} demand extremely low encoding and decoding latency, while robotics applications~\cite{li2024reality,han2025re3sim} require high inference throughput and minimal memory usage for onboard deployment. 

Besides, another major limitation of existing evaluations is the absence of geometric assessment. Geometric reconstruction is a central problem in computer graphics, and numerous recent works~\cite{huang20242d,chen2024pgsr,held2025tsplus,guo2024tetsphere} focus on improving the surface reconstruction quality of 3DGS, drawing substantial attention from the community.
However, research on 3DGS compression has so far concentrated almost exclusively on photorealistic metrics such as PSNR and SSIM. Although these metrics capture appearance fidelity, they overlook the fact that 3DGS is fundamentally a three-dimensional representation. As a result, existing evaluations fail to reveal how compression affects the geometric integrity of the underlying 3D structure.

To mitigate the aforementioned issues, we have developed a unified training and evaluation framework, characterized by the following features: \textbf{1) Standardization}. It defines a unified pipeline encompassing data loading, model training, compression encoding, and rendering evaluation, supporting the rapid development of new methods. \textbf{2) Decoupling}. We decouple components such as the rasterizer, entropy module, and codec as independent modules. This facilitates the combination of different methods and supports the integration of new modules in the future. \textbf{3) More Metrics}. Based on the same evaluation pipeline, we have not only included metrics such as PSNR but also extended to Gaussians count, peak memory usage, and geometric reconstruction accuracy, in order to provide a more comprehensive benchmark.

In summary, our contributions are threefold:

\begin{itemize}
    \item We propose a unified training and evaluation framework for 3DGS compression, which alleviates reproducibility issues of existing methods and lowers the development barrier for new methods.   
    \item We reproduce existing 3DGS compression methods under the proposed framework, providing the community with a fair, reliable, and comprehensive technical benchmark.
    \item Based on the framework, we have developed a new compression model, ChimeraGS, which has demonstrated competitive results across several dimensions, highlighting the flexibility and effectiveness of this framework.

\end{itemize}

\section{Related Work}
\label{sec:formatting}

The high rendering quality and flexibility of 3DGS have motivated research aimed at improving both reconstruction fidelity and storage efficiency. In this section, we review related works on 3D Gaussian Splatting and the methods proposed to compress and optimize the representation.

%-------------------------------------------------------------------------
\subsection{Novel View Synthesis}
Novel view synthesis is a classic problem in graphics research. Many work on the topic focused on image-based rendering \cite{wang2021ibrnet, flynn2024quark} and Neural Radiance Fields \cite{mildenhall2021nerf, barron2023zip}. In recent years, 
3D Gaussian Splatting (3DGS)~\cite{kerbl20233d} has emerged as an efficient and expressive representation for novel view synthesis and 3D reconstruction. By explicitly modeling scenes with anisotropic Gaussians, 3DGS achieves high-quality rendering and real-time performance, attracting extensive research attention. 

Numerous works have improved the rendering pipeline through adaptive density control~\cite{hanson2025speedy}, anti-aliasing~\cite{yu2024mip} and more realistic shading mechanisms~\cite{yao2024reflective}. These advances consolidate 3DGS as a versatile representation bridging explicit geometry and neural rendering. Our framework supports various static Gaussian Splatting pipelines with distinct models and optimization objectives, providing a unified platform for evaluation and analysis.

%--------------------------------------------- ----------------------------
\subsection{Compression of 3DGS}
Despite its rendering efficiency, 3DGS models are storage-intensive, prompting research on compact representations. Pruning-based methods\cite{hanson2025pup, ali2024trimming,zhang2025consistent,fan2024lightgaussian,hanson2025speedy} remove redundant or visually insignificant Gaussians by analyzing opacity, gradient importance, or visibility contribution, effectively reducing both model size and rendering cost. 
Entropy-based compression\cite{niedermayr2024compressed, chen2024hac, zhan2025cat} further reduces storage through learned probabilistic coding of Gaussian parameters or by applying rate–distortion optimization to balance bit usage and rendering accuracy. Although these studies have substantially advanced 3DGS compression, their implementations are often inconsistent, making cross-method evaluation difficult. To address this, our work introduces a unified framework that integrates representative pruning, quantization, distillation, and entropy coding techniques under a consistent training and evaluation pipeline, facilitating reproducible and comprehensive benchmarking for future research.

%--------------------------------------------- ----------------------------
\subsection{Benchmark and Tools}

Unlike fully neural network-based models, the 3DGS framework incorporates numerous components beyond conventional neural network architectures, with the rasterizer being a typical example. Furthermore, the inherently complex control flow of 3DGS has generated a need for substantial boilerplate code.

GSplat~\cite{ye2025gsplat} serves as an open-source, efficient, and user-friendly library that provides an efficient implementation of 3D Gaussian Splatting.
GauStudio~\cite{ye2024gaustudio} serves as a framework for
modeling 3D Gaussian Splatting to provide standardized, plug-and-play components for users to easily customize and implement a 3DGS pipeline.
GSCodec Studio~\cite{li2025gscodec} focuses specifically on Gaussian Splat compression, offering a modular design for integrating and benchmarking different compression strategies such as pruning, quantization, and entropy coding. Beyond tools,
3DGS.zip~\cite{bagdasarian20253dgs} introduces a comprehensive compression for 3DGS compression task, and compared the characteristics of various methods.

Our work is also inspired by these studies and possesses unique characteristics. The most representative aspect is the design of a flexible modular mechanism, which integrates over 10 different rasterizers and other modules including entropy estimation and encoding. Furthermore, based on a scheduler mechanism, we have implemented a more streamlined template for Gaussian Splatting  models. Using this foundation, we have reproduced several representative methods and conducted comprehensive performance comparisons.

\section{Architecture}
\begin{figure*}[t]
\centering
\includegraphics[width=\textwidth]{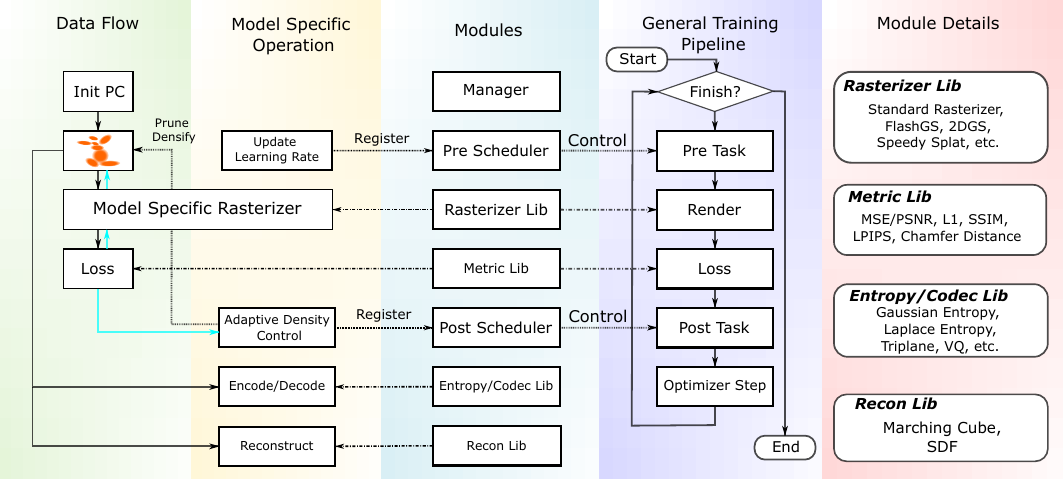}
\caption{Framework of Splatwizard. We have abstracted the Gaussian training process into a universal training pipeline, which can support various custom operations through a task scheduler mechanism. The horizontal arrow lines in the diagram illustrate the modules used at different stages of Gaussian training. For instance, different models employ different adaptive density control mechanisms, and such control mechanisms can be registered into the scheduler. This results in a GS model implemented based on Splatwizard, while also seamlessly leveraging various auxiliary modules and functions provided by the framework. }
\label{fig:framework of Splatwizard}
\vspace{-1em}
\end{figure*}
In Fig. \ref{fig:framework of Splatwizard}, we conceptualize our Splatwizard framework.
The primary objective of this framework is to provide a unified and reusable training and evaluation pipeline, thereby simplifying the implementation and research of future GS models. In the subsequent sections, we will provide a detailed introduction to the key mechanisms and functional modules of Splatwizard.
\subsection{Dynamic and AOT module loading}

A challenge in 3DGS research lies in the frequent need to modify or extend the rasterizer \cite{fan2024lightgaussian, huang20242d, feng2025flashgs, liao2025tc}. 
These modifications often tailored to specific tasks and
introduce a host of non-research-related hurdles that hinder productivity. 
This problem is further compounded in compression-focused 3DGS tasks, where the integration of diverse entropy encoders \cite{chen2024hac, zhan2025cat}
adds another layer of complexity. These encoders, often implemented in  C++ or CUDA for efficiency, introduce additional build dependencies, versioning conflicts, and compatibility issues with existing rasterizer codebases. As a result, researchers spending significant time resolving linker errors, adjusting makefiles, or ensuring cross-platform consistency—efforts that contribute little to the scientific goals of their work.

To address these challenges, we designed a flexible module system leveraging PyTorch’s built-in loading mechanisms and extension frameworks \cite{paszke2019pytorch}. This system supports both Just-in-Time (JIT) loading and Ahead-of-Time (AOT) compilation, enabling seamless integration of custom components with minimal configuration. For JIT loading, modules are compiled dynamically at runtime using PyTorch’s C++ extension APIs. For AOT compilation, modules are built during installation, generating platform-specific binaries that can be loaded instantly at runtime.

Built on this foundation, we have integrated distinct rasterizers and a suite of CUDA/C++ modules. Critically, this system abstracts all compilation details from end users: regardless of whether a module is JIT-loaded or pre-built, researchers can import and use it via simple Python import. The underlying framework handles dependency resolution, and kernel loading, ensuring that even complex combinations of rasterizers and compression modules work out-of-the-box.

\subsection{Unified Pipeline}
Another characteristic of 3DGS is that its training process involves many operations beyond parameter updates and gradient computations \cite{kerbl20233d, lee2024compact}. The most representative of these are the pruning and densification of Gaussian points. Additionally, many methods incorporate operations that are coupled with training epochs. For example, in HAC \cite{chen2024hac}, the training is mainly divided into full-precision training, training with simulated quantization, and training with entropy constraints. Beyond these three main training stages, there are also brief pauses in pruning/densification operations. The training phases in CAT-3DGS \cite{zhan2025cat} are even more complex. Manually controlling these intricate training processes solely through if-else code is both cumbersome and error-prone. Therefore, we have implemented a unified training workflow based on a scheduler mechanism. A general flowchart can be referred to in Fig \ref{fig:framework of Splatwizard}.

\begin{listing}[t]
 
\begin{minted}[mathescape,
               linenos,
               numbersep=5pt,
               fontsize=\footnotesize,
               gobble=0]{python}
class GaussainModel:
  def register_pre_task(
    self, 
    scheduler: Scheduler, 
    ppl: PipelineParams, 
    opt: OptimizationParams
  ):
    scheduler.register_task(
      range(opt.iterations), 
      task=self.update_learning_rate
    )   
\end{minted}
% \vspace{-1em}
\caption{Snippet used to register task.}
\label{register_task}
\end{listing}

We have unified the training process of 3DGS into five stages: pre-scheduler execution, rendering, loss calculation, post-scheduler execution, and optimizer update. The scheduler is designed to execute specified tasks at designated training iterations, and Listing \ref{register_task} demonstrates the API for task registration in the scheduler. When the scheduler executes tasks, it automatically uses appropriate parameters for dispatch based on the type annotations \cite{di2022evolution} of the task's formal parameters. With this design, various operations beyond rendering and loss calculation can be converted into tasks and clearly presented through the scheduling flow. Taking the training process of the original Gaussian as an example, Table \ref{tab:3dgs_schedule_plan} shows the tasks included in the pre-scheduler and post-scheduler. One constraint in task allocation is that tasks requiring the use of rendering results must be registered in the post-scheduler, as the pre-scheduler cannot access rendering results.

In addition to clarifying the training workflow, another advantage of the scheduler-based mechanism is its similarity in code organization to the original 3DGS implementation. This allows for easy adaptation of various new works based on the original codebase into a scheduler-based organizational structure. This significantly reduces the difficulty of integrating new methods into our framework.

\begin{table}[t]
    \centering
    \resizebox{0.95\linewidth}{!}{
    \begin{tabular}{c c c}
    \toprule
         Stage  & Task & Plan   \\
         \midrule
         Pre  & update lr & range(30000) \\
          & increase SH degree & range(0, 30000, 1000) \\
          \midrule
          Post & collect gradient statistics & range(15000) \\
          & prune and densify & range(500, 15000, 100) \\
          & reset opacity & range(0, 15000, 3000) \\
         \bottomrule
    \end{tabular}}
    \caption{Schedule plan for vanilla 3DGS. We use python style range expression to illustrate detail execution time.}
    \label{tab:3dgs_schedule_plan}
    \vspace{-2em}
\end{table}

\subsection{Function Library}

The Rasterizer Lib is a collection of encapsulated rasterizer from previous works.
This includes a standard rasterizer~\cite{kerbl20233d}, along with its subsequent accelerated version~\cite{mallick2024taming} and an additional variant that incorporates depth computation~\cite{kerbl2024hierarchical}. Since pruning methods often rely on importance assessment~\cite{hanson2025pup}, the framework provides rasterizers capable of calculating global importance scores based on opacity~\cite{fan2024lightgaussian}, as well as variants that support computing reparameterized approximate Hessian importance~\cite{hanson2025speedy}. Additionally, it includes versions designed specifically for reconstruction, such as 2DGS~\cite{huang20242d} and TrimGS ~\cite{fan2024trim}. The architectural design of Splatwizard allows for flexible combination of these modules, enabling the mixed use of different rasterizers within a single method.

The Metrics Lib module focuses on loss calculation and performance metric evaluation, providing rich tool support for the Loss module. It includes the implementation of various loss functions  and the calculation of image quality metrics (PSNR, SSIM~\cite{wang2004image} and LPIPS~\cite{zhang2018unreasonable}). In addition to these image quality assessment metrics, we have specifically introduced geometric evaluation metrics, including Chamfer distance of point cloud~\cite{fan2024trim} and reconstructed mesh~\cite{huang20242d}. While such metrics are often unreported in previous compression models, we believe their inclusion can provide a more comprehensive reflection of model performance.

We have categorized modules related to entropy estimation, quantization, and entropy encoding into the Codec Lib. Current research on 3DGS compression has applied a variety of entropy models and encoding algorithms. In terms of quantization, methods such as STE quantization, noise relaxation quantization, vector quantization, and training-aware quantization are included. For context modeling, models based on hash grids and triplanes have also been utilized. As for entropy encoders, techniques such as arithmetic coding, rANS coding, and Huffman coding are covered. These diverse methods were previously scattered across different studies, and consolidating these modules here will facilitate future research.

Reconstruction Lib acts as a critical bridge between 3DGS and mesh-based  models, converting optimized Gaussian primitives into structured meshes. Following previous work, we render depth maps of the training views and utilize truncated signed distance fusion (TSDF) to fuse the reconstruction depth maps, using Open3D~\cite{zhou2018open3d}.

\subsection{Model Zoo}
Building on the aforementioned modular framework, we have systematically reproduced several mainstream algorithms in 3DGS compression, forming a unified benchmark implementation. Throughout the reproduction process, we adhered principle of minimizing reproduction discrepancies: on one hand, we precisely aligned key parameter settings described in the original papers to ensure that the critical logic of each algorithm remains consistent with the original descriptions; on the other hand, leveraging the framework's modular design, we abstracted common components across different algorithms into shared modules, only implementing unique innovations of each algorithm. This approach reduce performance deviations caused by differences in engineering implementation.
Notably, during the reproduction process, we identified and fixed several potential issues in the original code that affected result consistency. 
Through this approach, our implementation not only provides a unified comparison benchmark for 3DGS compression algorithms but also offers a more robust code reference for the community, effectively reducing discrepancies in research caused by differences in engineering implementations.

\section{ChimeraGS}
\begin{figure*}[t]
\centering
\includegraphics[width=\textwidth]{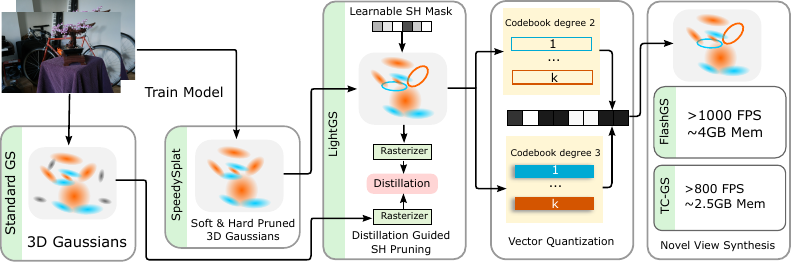}
\caption{Framework of ChimeraGS. Since Splatwizard natively supports flexible module composition, we can easily mix multiple different rasterizers in same model. The labels on the left side indicate the rasterizers used in each phase.}
\label{fig:framework_chimera}
\vspace{-1em}
\end{figure*}
As a demonstration, we have developed a new model with closer comprehensive performance comparing with end-to-end trained model like HAC~\cite{chen2024hac} by strategically integrating effective modules and proposed distillation guided SH pruning. Fig. \ref{fig:framework_chimera} illustrate the framework of our method.

\subsection{Pruning Strategy}
The original LightGaussian~\cite{fan2024lightgaussian} comprises multiple stages: pruning, distillation, and the final vector quantization encoding phase. Each of these stages has room for optimization. During the pruning stage, we can adopt more aggressive pruning methods and ratios to reduce the number of Gaussian primitives. For example, we can adopt the pruning strategy from PUP-3DGS~\cite{hanson2025pup} to significantly reduce the number of Gaussian kernels after pruning while maintaining comparable rendering quality. Additionally, beyond post-processing pruning, we can directly learn a compact Gaussian representation to replace the pruning stage. This approach often achieves higher compression rates.

We adopt the SpeedySplat~\cite{hanson2025speedy} to obtain a highly compact initialization of Gaussians. SpeedySplat utilizes the reparameterized approximation of Hessian matrix to measure the importance of each Gaussian primitive 
\begin{equation}
    \tilde{U}_i = \log |\nabla_{g_i} I_\mathcal{G} \nabla_{g_i} I_\mathcal{G}^T|,
    \label{speedy_app}
\end{equation}
where $I_\mathcal{G}$ is rendered image. $g_i$ is the 2D projected value of $G_i$ at pixel $p$, given by
\begin{equation}
    g_i = e^q, q = -\frac{1}{2}(p - \mu_{i_{2D}})\boldsymbol{\Sigma}_{i_{2D}}^{-1}(p - \mu_{i_{2D}})^T.
\end{equation}
Since $\log$ is monotonically increasing, Equation \ref{speedy_app} can be further rewrited as
\begin{equation}
    \tilde{U}_i = (\nabla_{g_i} I_\mathcal{G})^2.
    \label{speedy_app}
\end{equation}
Based on this efficient importance evaluation metric, effective pruning was achieved during the training phase.

\subsection{Distillation Guided SH Pruning}
LightGaussian further reduces the number of parameters per Gaussian primitive by directly reduce the spherical harmonic degree and then applying distillation fine-tuning. This approach significantly reduces the encoded size. However, adopting a more aggressive strategy during the pruning stage may result in a scenario where simply reducing the order followed by fine-tuning fails to adequately compensate for the quality loss caused by the degree reduction. Therefore, we have adopted a dynamic pruning approach here, which determines which Gaussian primitives undergo order reduction in a learnable manner.

More specifically, We apply binary masks $M\in\{0, 1\}^N$ to degree-3 SH parameter of each Gaussian primitive. 
\begin{equation}
    \hat{\mathbf{c}}_n^{(3)} = M_n \mathbf{c}_n^{(3)},
\end{equation}
where $n$ is the index of the primitive, $\mathbf{c}_n^{(3)}$ is degree-3 SH parameter. During rendering, we use $\hat{\mathbf{c}}_n^{(3)}$ instead of $\mathbf{c}_n^{(3)}$.

Obviously, binary mask $M$ is non-differentiable, following previous works, we use a straight-through-estimator
\begin{equation}
    M_n = \operatorname{sg}(\mathbbm{1}[\sigma(m_n) > c] - \sigma(m_n)) + \sigma(m_n), 
\end{equation}
where $\operatorname{sg}(\cdot)$ is stop gradient operator, $\mathbbm{1}[\cdot]$ and $\sigma(\cdot)$ are indicator function and sigmoid respectively. To eliminate redundant degree-3 SH parameters, an additional loss is required
\begin{equation}
    \mathcal{L}_m = \frac{1}{N} \sum_{n=1}^{N} \sigma(m_n).
\end{equation}

The use of learnable masks for pruning Gaussians or Gaussian parameters has also been explored in previous works~\cite{lee2024compact, wang2024end}. Our main improvement here is to leverage distillation to further enhance performance. Following LightGaussian~\cite{fan2024lightgaussian},
we use a teacher model to generate pseudo view to supervise student model
\begin{equation}
\mathcal{L}_{\text{distill}} = \frac{1}{HW}\left\| I_{\text{teacher}}(\mathbf{R}|\mathbf{t}) - I_{\text{student}}( \mathbf{R}|\mathbf{t}) \right\|_2^2,
\end{equation}
where $I_{\text{teacher}}(\mathbf{R}|\mathbf{t})$ and $I_{\text{student}}( \mathbf{R}|\mathbf{t})$ are the rendered image given a camera rotation $\mathbf{R}$ and position $\mathbf{t}$ of teacher model and student model respectively. The position is given by
\begin{equation}
\mathbf{t}_{\text{pseudo}} = \mathbf{t}_{\text{train}} + \mathcal{N}(0, \sigma^2),
\end{equation}
where $\mathbf{t}_{\text{train}}$ is sampled position, $\mathcal{N}$ is denotes a Gaussian distribution with mean 0 and variance $\sigma^2$. The final loss in distillation stage is
\begin{equation}
    \mathcal{L} = \mathcal{L}_{\text{distill}} + \lambda  \mathcal{L}_m 
\end{equation}

By comprehensively applying learnable masks and knowledge distillation, our method can mitigate the performance loss caused by pruning, thereby improving the rate-distortion performance of the model.

\subsection{Encoding and Rendering}

In the encoding phase, we adopt the vector quantization encoder from LightGaussian. The main difference is that we use separate codebooks for the degree-2 and degree-3 SH parameters respectively. Since Gaussian rendering is position-order-independent, Gaussian points can be encoded separately according to their SH degrees. Additionally, to further improve encoding performance, we sort the Gaussian points using Morton order before the final encoding.

Since our method does not rely on a specially implemented rasterizer during the rendering phase, it can be easily integrated with existing rendering acceleration schemes. Here, we have experimented with FlashGS~\cite{feng2025flashgs} and TC-GS~\cite{liao2025tc} respectively. Combined with the model size advantage brought by compression itself, both methods have achieved impressive rendering speed.

\section{Benchmarking}
\label{sec:experiments}

\begin{figure*}
\centering
\includegraphics[width=\textwidth]{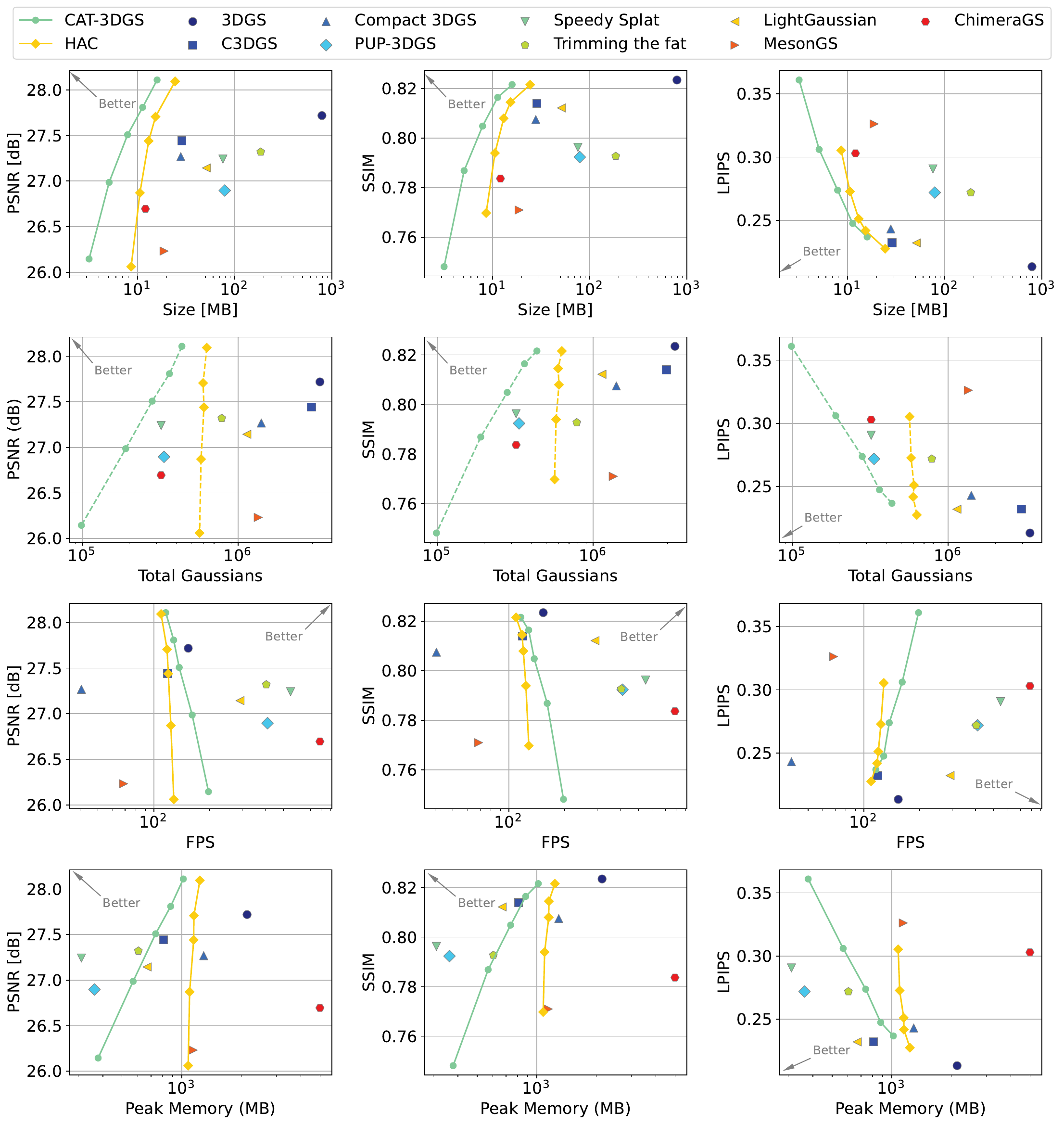}
\caption{Results of Mip-NeRF 360. Note that when counting the total number of Gaussian primitives in HAC and CAT-3DGS, only the number of anchor is considered, and the values are not directly comparable to other methods, hence they are drawn with a dashed line.}
\label{fig:mipnerf_results}
\vspace{-1em}
\end{figure*}
\subsection{Benchmarking Setup}
\label{subsec:experimental_setup}

Our evaluation is conducted on six multi-view image datasets: Mip-NeRF 360~\cite{barron2022mipnerf360}, Tanks \& Temples~\cite{Knapitsch2017}, Deep Blending~\cite{10.1145/3272127.3275084}, BungeeNeRF~\cite{xiangli2022bungeenerf}, NeRF Synthetic~\cite{mildenhall2020nerf} and DTU. Tanks\&Temples features real-world scenes with complex geometry captured using laser scanners, while Mip-NeRF 360 contains 9 unbounded scenes enabling 360-degree rendering. Deep Blending provides diverse indoor and outdoor environments, BungeeNeRF focuses on extreme multi-scale scenarios, and NeRF Synthetic offers 8 photorealistic Blender-rendered scenes. For geometric evaluation, DTU dataset \cite{jensen2014large} provides measure based geometry ground truth. This enables us to evaluate the reconstruction accuracy of the model, thereby providing a more comprehensive reflection of its performance.

In terms of model selection, we evaluated all methods in the model zoo, with specific methods detailed in the Table \ref{tab:intro:comparision}. All approaches were implemented or reproduced based on SW, using unified evaluation protocols and consistent training/testing dataset splits. Furthermore, to prevent potential data leakage from affecting bitrate evaluation, we strictly isolated the models used during the encoding and decoding phases. All included methods support encoding complete information into a single binary file, while the decoding phase utilizes only a newly initialized model and the corresponding binary file for testing.

\subsection{Photorealistic Metrics Results}

\begin{figure*}[ht]
\centering
\includegraphics[width=0.95\textwidth]{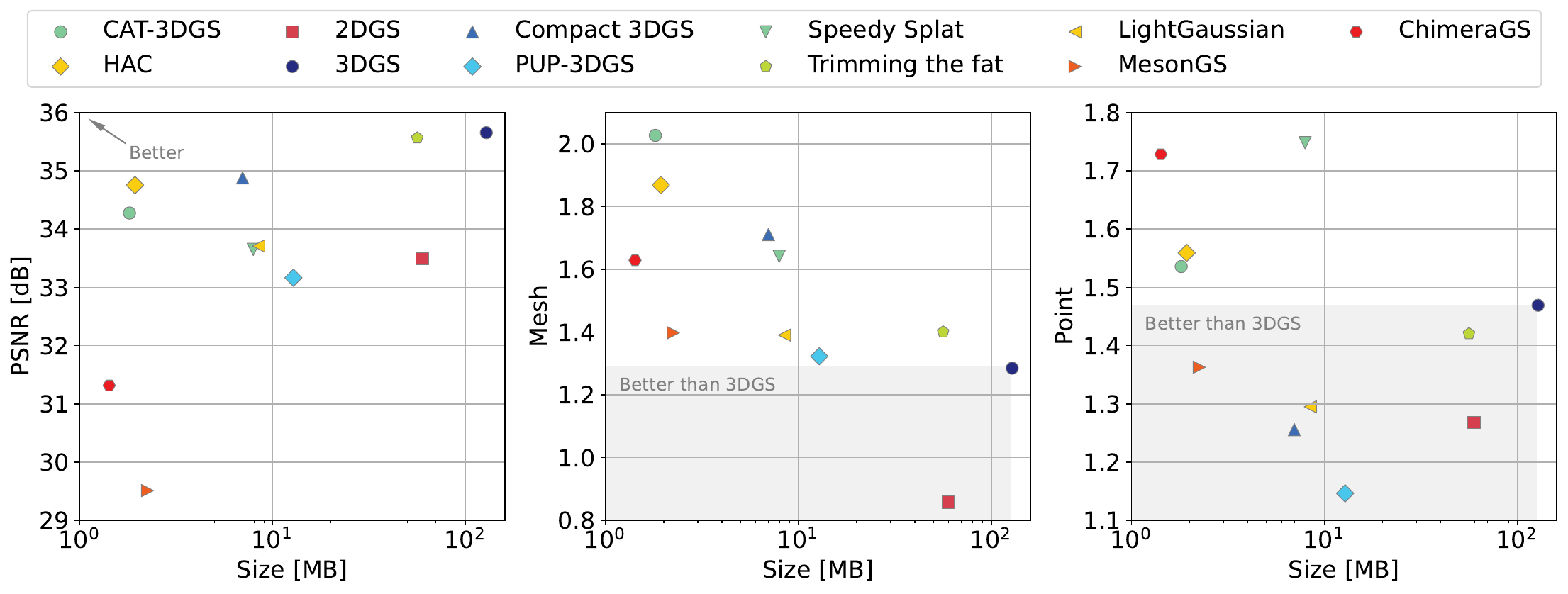}
\caption{Results of DTU dataset. Here we present the PSNR, along with the Chamfer distance based on mesh and point cloud respectively.}
\label{fig:geometric_results}
\vspace{-1em}
\end{figure*}

 Fig. \ref{fig:mipnerf_results}  presents evaluation results of all methods on MipNeRF, with results on additional datasets included in the Supplementary. Beyond the previously mentioned PSNR, SSIM, and LPIPS metrics, we also measured bitrate, Gaussian count, inference rendering speed, and peak memory usage during rendering. The figure reports cross-comparison results of these metrics.

 From the conventional rate-distortion curves, we observe a trend: methods employing hybrid implicit neural networks and 3DGS generally achieve superior rate-distortion performance, while models that combine quantization operations with rate-distortion optimization also demonstrate relatively competitive performance. However, when considering Gaussian count as the metric for bitrate, the aforementioned observations may not necessarily hold.

Furthermore, comparisons of rendering frame rate and peak GPU memory usage provide a more comprehensive perspective. As expected, hybrid representation-based models generally exhibit lower frame rates compared to pruning-based approaches, though their impact on memory consumption is not significant.

The proposed ChimeraGS also achieves competitive results across several dimensions. Compared to prior approaches, our method attains rate-distortion performance closer to HAC \cite{chen2024hac}, while achieving top-tier results in terms of Gaussian count. However, experimental observations reveal that multi-stage models like LightGaussian \cite{fan2024lightgaussian} incur additional performance overhead between stages. Combined with previous findings, this insight points toward future improvement directions, such as incorporating quantization during training to enhance performance.

\subsection{Geometric Metrics Results}
Fig. \ref{fig:geometric_results} shows the rate-distortion performance of geometric metrics. Unlike the previous section, the evaluation for photo employs the novel view synthesis task, where the dataset is divided into training and test sets, and the decoded model is only tested on the test set for reconstruction metrics. In the measurement of geometric metrics, we primarily focus on geometric reconstruction accuracy; therefore, the entire dataset is used for training. Correspondingly, the rate-distortion curves here only represent data fitting capability, not novel view synthesis capability. Since current TSDF-based mesh reconstruction methods rely on depth maps derived from Gaussians, for all models using standard rasterizers (including hybrid models), we have replaced their rasterizers with versions that support depth computation to facilitate subsequent mesh reconstruction stages.

Compared to photo-level evaluation, we find that some methods even achieve point cloud-level accuracy that approaches or surpasses the performance of 2DGS. This suggests that compression tasks and geometric reconstruction tasks are not entirely conflicting objectives. In the future, we could explore integrating methods from these two specialized tasks to investigate pathways for enhancing each other's performance. Meanwhile, we should also note that not every compressor can improve geometric quality, so the design of specific methods remains particularly important.

\section{Conclusion and Future Work}
In this paper, we introduce a universal framework for GS compression evaluation, named Splatwizard. This framework provides a unified training and evaluation pipeline adaptable to various Gaussian methods, while offering a flexible module utilization mechanism and a comprehensive function library. Based on this framework, we have reproduced multiple Gaussian compression methods and, leveraging the flexible architecture, developed a distinctive approach with unique performance characteristics. Experimental results include multi-dimensional comparisons of all methods.
% using geometric metrics and other evaluation criteria. 
We hope this work will accelerate research in this field by providing fundamental tools and resources.
In the future, we will further expand the capabilities of this framework by incorporating a wider range of methods for comparison, while systematically organizing its functional modules to facilitate their use in subsequent work. Additionally, while the current framework is primarily tailored for static GS scene compression methods, we plan to extend its functionality to tasks like dynamic scenes~\cite{yang2024deformable, wu20244d}, feed-forward paradigm~\cite{chen2024fast} and streaming 3DGS compression~\cite{sun20243dgstream}.
{
    \small
    \bibliographystyle{ieeenat_fullname}
    % \bibliography{main}

}

\clearpage
\setcounter{page}{1}
\maketitlesupplementary

% \section{Rationale}
% \label{sec:rationale}
% % 
% Having the supplementary compiled together with the main paper means that:
% % 
% \begin{itemize}
% \item The supplementary can back-reference sections of the main paper, for example, we can refer to \cref{sec:intro};
% \item The main paper can forward reference sub-sections within the supplementary explicitly (e.g. referring to a particular experiment); 
% \item When submitted to arXiv, the supplementary will already included at the end of the paper.
% \end{itemize}
% % 
% To split the supplementary pages from the main paper, you can use \href{https://support.apple.com/en-ca/guide/preview/prvw11793/mac#:~:text=Delete%20a%20page%20from%20a,or%20choose%20Edit%20%3E%20Delete).}{Preview (on macOS)}, \href{https://www.adobe.com/acrobat/how-to/delete-pages-from-pdf.html#:~:text=Choose%20%E2%80%9CTools%E2%80%9D%20%3E%20%E2%80%9COrganize,or%20pages%20from%20the%20file.}{Adobe Acrobat} (on all OSs), as well as \href{https://superuser.com/questions/517986/is-it-possible-to-delete-some-pages-of-a-pdf-document}{command line tools}.

\section{Schedule plan of methods}
In this section, we will introduce the execution plans for all methods included. We use python style range expression to illustrate detail execution time.

\subsection{3DGS}
\begin{table}[h]
    \centering
    \resizebox{0.95\linewidth}{!}{
    \begin{tabular}{c c c}
    \toprule
         Stage  & Task & Plan   \\
         \midrule
         Pre  & update lr & range(30000) \\
          & increase SH degree & range(0, 30000, 1000) \\
          \midrule
          Post & collect gradient statistics & range(15000) \\
          & prune and densify & range(500, 15000, 100) \\
          & reset opacity & range(0, 15000, 3000) \\
         \bottomrule
    \end{tabular}}
    \caption{Schedule plan for vanilla 3DGS. }
    \label{tab:3dgs_schedule_plan}
    \vspace{-1em}
\end{table}

\subsection{Trimming the Fat}
\begin{table}[h]
    \centering
    % \resizebox{0.95\linewidth}{!}{
    \begin{tabular}{c c c}
    \toprule
         Stage  & Task & Plan   \\
         \midrule
         Pre  & update lr & range(30000) \\
          \midrule
          Post & gradient aware prune & range(100, 5000, 500) \\

         \bottomrule
    \end{tabular}
    % }
    \caption{Schedule plan for Trimming the Fat.}
    \label{tab:3dgs_schedule_plan}
\vspace{-1em}
\end{table}

\subsection{PUP-3DGS}
\begin{table}[h]
    \centering
    % \resizebox{0.95\linewidth}{!}{
    \begin{tabular}{c c c}
    \toprule
         Stage  & Task & Plan   \\
         \midrule
         Pre  & update lr & range(30000) \\
          \midrule
          Post & gradient aware prune & range(100, 5000, 500) \\

         \bottomrule
    \end{tabular}
    % }
    \caption{Schedule plan for PUP-3DGS.}
    \label{tab:ttf_schedule_plan}

\end{table}

% \begin{table}[h]
%     \centering

%     \begin{tabular}{c c c}
%     \toprule
%          Stage  & Task & Plan   \\
%          \midrule
%           Post & prune & [1, 5001] \\

%          \bottomrule
%     \end{tabular}
%     \caption{Schedule plan for PUP-3DGS.}
%     \label{tab:pup_schedule_plan}

% \end{table}

\subsection{Speedy-Splat}
\begin{table}[h]
    \centering
    \resizebox{0.95\linewidth}{!}{
    \begin{tabular}{c c c}
    \toprule
         Stage  & Task & Plan   \\
         \midrule
          Pre  & update lr & range(30000) \\
          & increase SH degree & range(0, 30000, 1000) \\
          \midrule
          Post & collect gradient statistics & range(15000) \\
          & prune and densify & range(500, 15000, 100) \\
          & reset opacity & range(0, 15000, 3000) \\
& soft prune  & range(6000, 15000, 3000) \\
& hard prune & range(15000, 30000, 3000) \\
         \bottomrule
    \end{tabular}}
    \caption{Schedule plan for Speedy-Splat.}
    \label{tab:speedy_schedule_plan}

\end{table}

\subsection{LightGaussian}
Note we separate prune and distillation to different stages.
\begin{table}[h]
    \centering
    % \resizebox{0.95\linewidth}{!}{
    \begin{tabular}{c c c}
    \toprule
         Stage  & Task & Plan   \\

          \midrule
          Post & prune & 1 \\

         \bottomrule
    \end{tabular}
    % }
    \caption{Schedule plan for LightGaussian PRUNE stage.}
    \label{tab:lightgs_schedule_plan}

\end{table}

\begin{table}[h]
    \centering
    % \resizebox{0.95\linewidth}{!}{
    \begin{tabular}{c c c}
    \toprule
         Stage  & Task & Plan   \\

          \midrule
          Post & calculate importance score & 5000 \\

         \bottomrule
    \end{tabular}
    % }
    \caption{Schedule plan for LightGaussian DISTILL stage.}
    \label{tab:lightgs_d_schedule_plan}

\end{table}

\subsection{MesonGS}
\begin{table}[h]
    \centering
    % \resizebox{0.95\linewidth}{!}{
    \begin{tabular}{c c c}
    \toprule
         Stage  & Task & Plan   \\

          \midrule
          Pre & update lr & range(3000) \\
          & calculate importance score  & 1 \\
          & create octree  & 1 \\
         \bottomrule
    \end{tabular}
    % }
    \caption{Schedule plan for MesonGS.}
    \label{tab:lightgs_d_schedule_plan}

\end{table}

\subsection{C3DGS}

\begin{table}[h]
    \centering
    % \resizebox{0.95\linewidth}{!}{
    \begin{tabular}{c c c}
    \toprule
         Stage  & Task & Plan   \\

          \midrule
          Pre & update lr & range(30000) \\
          & VQ compress  & 1 \\
         \bottomrule
    \end{tabular}
    % }
    \caption{Schedule plan for C3DGS.}
    \label{tab:lightgs_d_schedule_plan}

\end{table}

\subsection{Compact3DGS}

\begin{table}[h]
    \centering
    \resizebox{0.95\linewidth}{!}{
    \begin{tabular}{c c c}
    \toprule
         Stage  & Task & Plan   \\
         \midrule
          Pre  & update lr & range(30000) \\
          & increase SH degree & range(0, 30000, 1000) \\
          & switch to RVQ training & 29000 \\
          \midrule
          Post & collect gradient statistics & range(15000) \\
          & prune and densify & range(500, 15000, 100) \\
          & reset opacity & range(0, 15000, 3000) \\
& mask prune  & range(15000, 30000, 1000) \\

         \bottomrule
    \end{tabular}}
    \caption{Schedule plan for Compact3DGS.}
    \label{tab:compact_schedule_plan}

\end{table}

\subsection{HAC}
\begin{table}[h]
    \centering
    \resizebox{0.95\linewidth}{!}{
    \begin{tabular}{c c c}
    \toprule
         Stage  & Task & Plan   \\
         \midrule
          Pre  & update lr & range(30000) \\
          & update anchor bound & 1 \\
          
          \midrule
          Post & collect gradient statistics & range(15000) \\
          & adjust anchor & range(1500, 3000, 100) \\
          & adjust anchor & range(4000, 15000, 100) \\
          & switch to quantized training & 3000 \\
          & switch to entropy training & 10001 \\ 

         \bottomrule
    \end{tabular}}
    \caption{Schedule plan for HAC.}
    \label{tab:compact_schedule_plan}

\end{table}

\subsection{CAT-3DGS}

\begin{table}[h]
    \centering
    \resizebox{0.95\linewidth}{!}{
    \begin{tabular}{c c c}
    \toprule
         Stage  & Task & Plan   \\
         \midrule
          Pre  & update lr & range(30000) \\
           & update cam mask & range(30000) \\
          & update anchor bound & 1 \\
          & setup triplane & 10000 \\
          & switch to quantized training & 3000 \\
          & switch to entropy training & 10001 \\ 
          
          \midrule
          Post & collect gradient statistics & range(15000) \\
          & adjust anchor & range(1500, 3000, 100) \\
          & adjust anchor & range(4000, 15000, 100) \\

         \bottomrule
    \end{tabular}}
    \caption{Schedule plan for CAT-3DGS.}
    \label{tab:compact_schedule_plan}

\end{table}

\section{More experiments results.}
These figures respectively show the results on the Deep Blending, NeRF Synthetic,  and BungeeNeRF datasets. All training and testing were completed on a single RTX 3090.  Note that when counting the total number of Gaussian primitives in HAC and CAT-3DGS, only the number of anchor is considered, and the values are not directly comparable to other methods, hence they are drawn with a dashed line.

\begin{figure*}
\centering
\includegraphics[width=\textwidth]{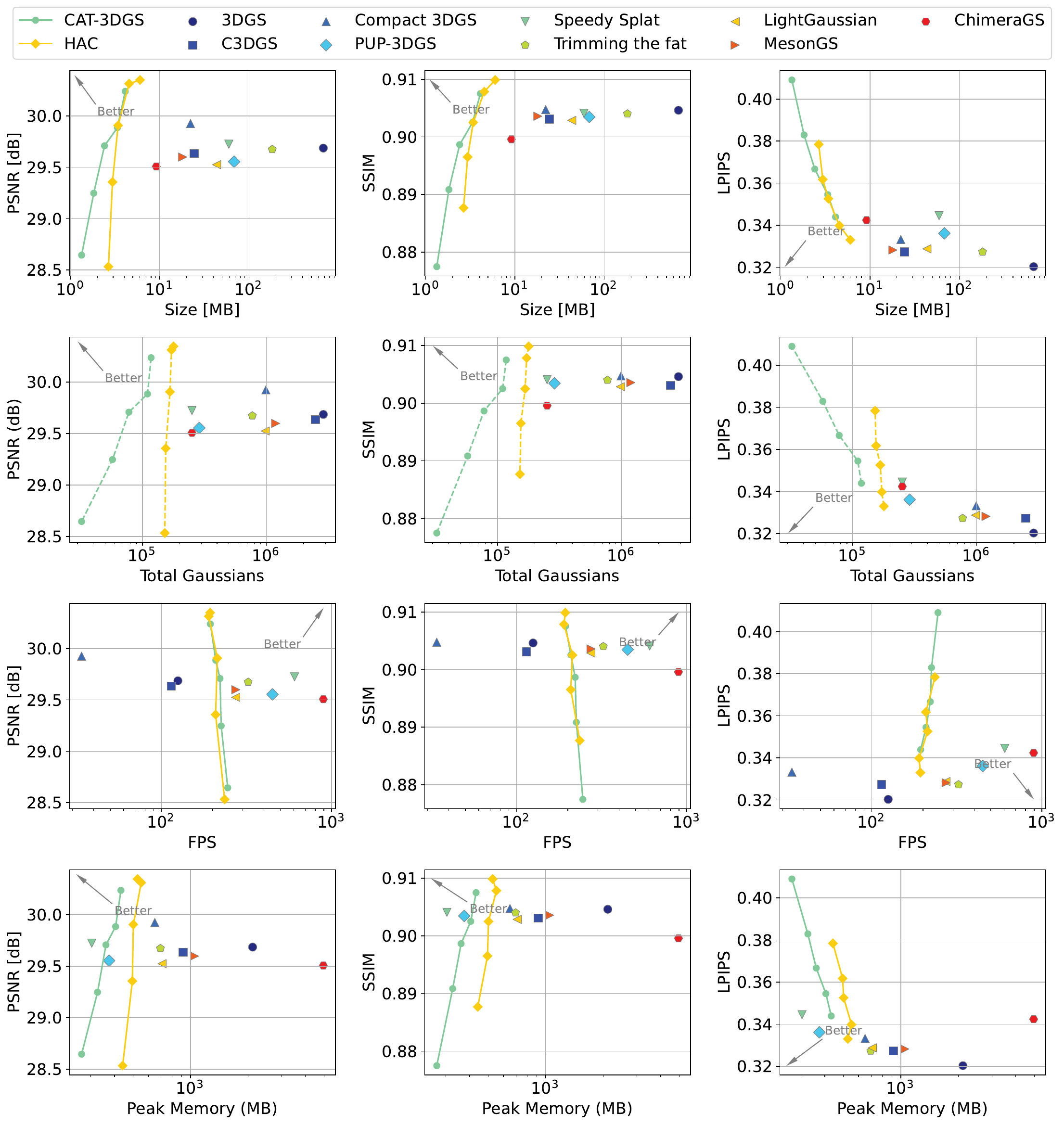}
\caption{Results of DeepBlending.}
\label{fig:db_results}
% \vspace{-1em}
\end{figure*}

\begin{figure*}
\centering
\includegraphics[width=\textwidth]{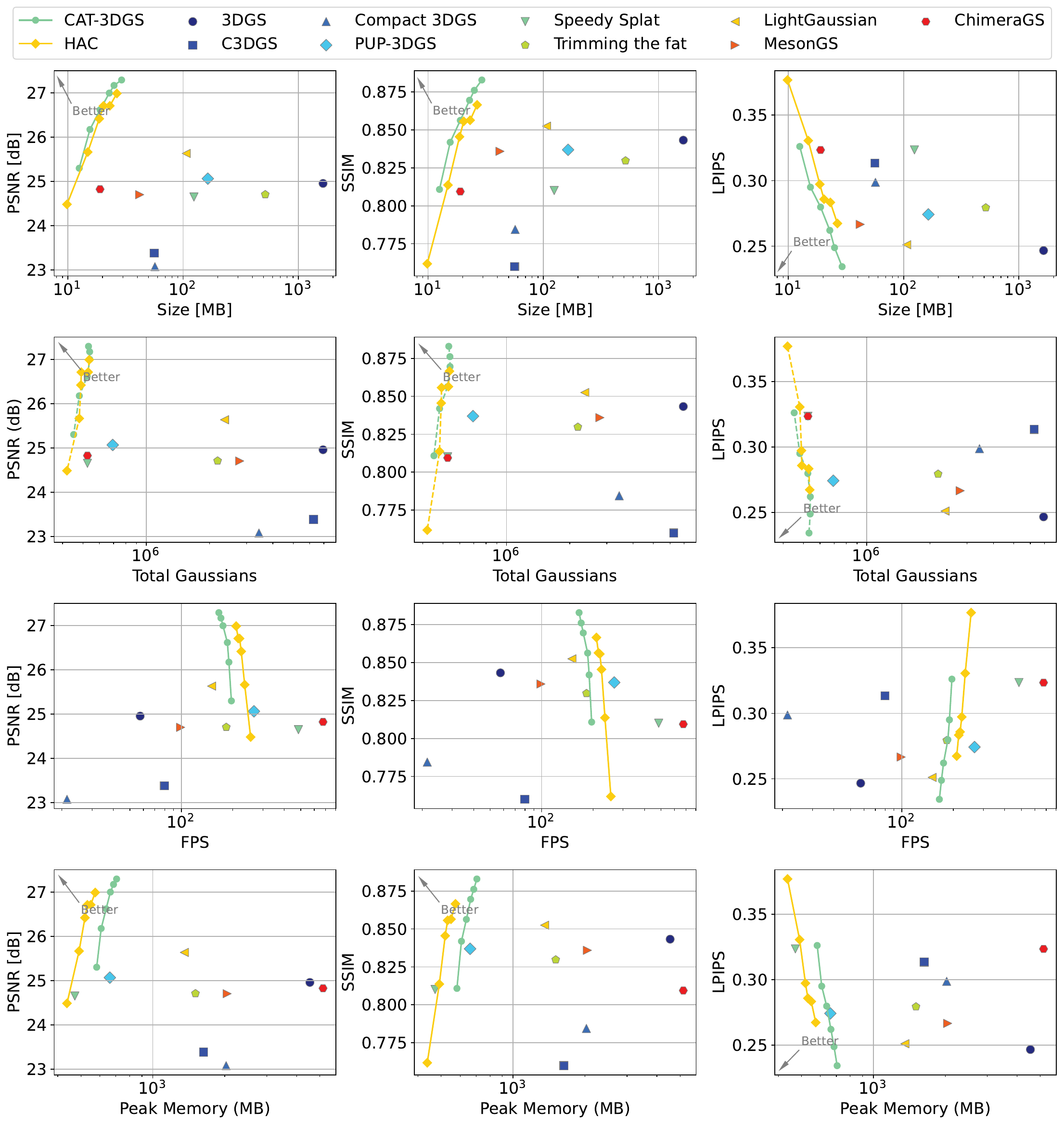}
\caption{Results of BungeeNeRF.}
\label{fig:bungee_results}
% \vspace{-1em}
\end{figure*}

\begin{figure*}
\centering
\includegraphics[width=\textwidth]{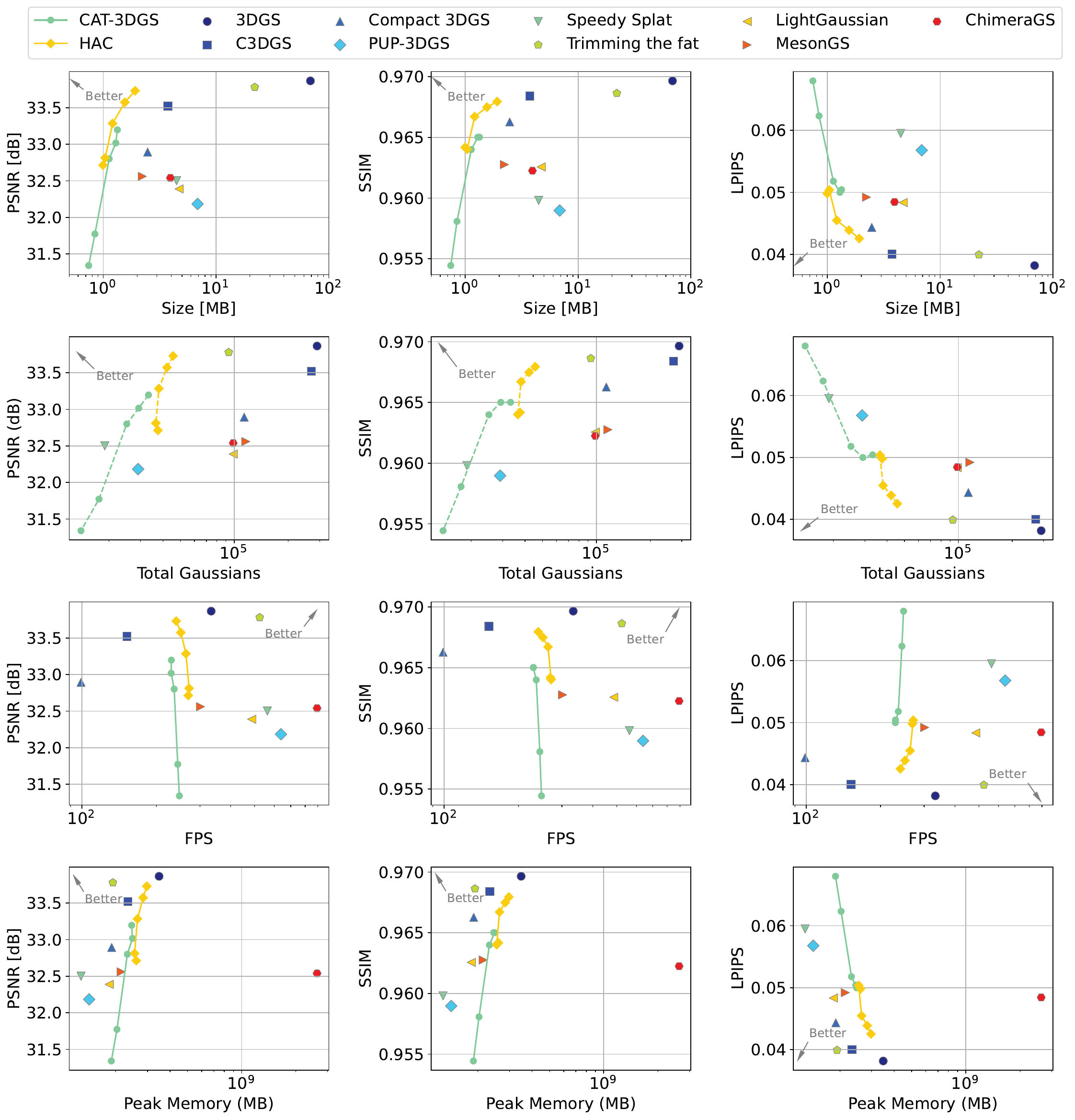}
\caption{Results of NeRF Synthetic.}
\label{fig:nerf_results}
% \vspace{-1em}
\end{figure*}

\begin{figure*}
\centering
\includegraphics[width=\textwidth]{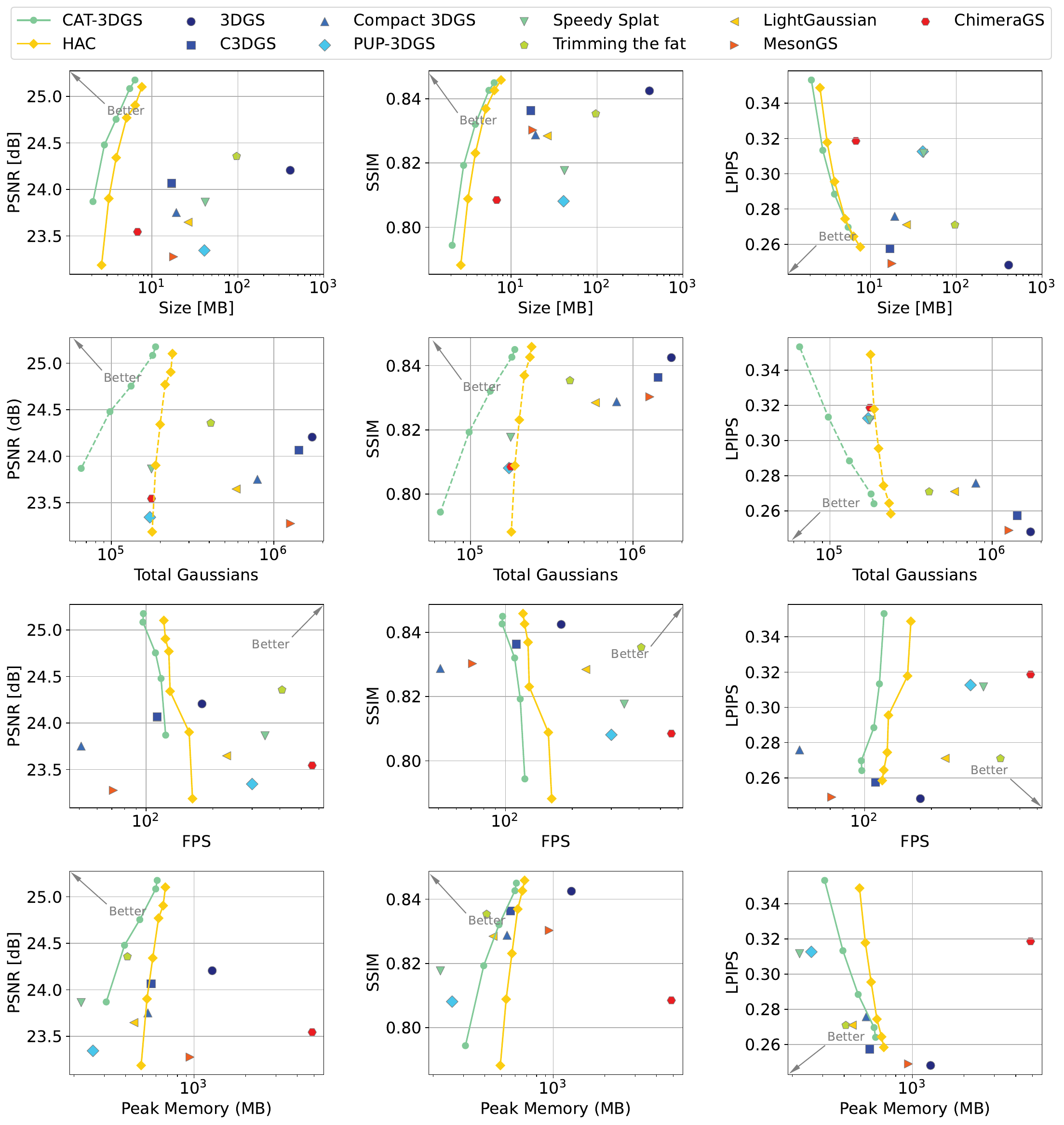}
\caption{Results of Tanks \& Temples.}
\label{fig:tt_results}
% \vspace{-1em}
\end{figure*}

\end{document}